\documentclass{article}

\usepackage{microtype}
\usepackage{graphicx}
\usepackage{subfigure}
\usepackage{caption}
\usepackage{multirow}
\usepackage{tablefootnote}
\usepackage{algorithm}
\usepackage{algorithmicx}
\usepackage{algpseudocode}
\usepackage{booktabs} 
\usepackage[dvipsnames]{xcolor}
\usepackage{hyperref}

\newcommand{\etal}{\textit{et al.}}
\usepackage{xspace}
\newcommand{\ourtitle}{NNSplitter\xspace}
\newcommand{\ouralg}{NNSplitter\xspace}

\usepackage[accepted]{icml2023}

\usepackage{amsmath}
\usepackage{amssymb}
\usepackage{mathtools}
\usepackage{amsthm}

\usepackage[capitalize,noabbrev]{cleveref}

\theoremstyle{plain}

\theoremstyle{definition}

\theoremstyle{remark}

\usepackage[textsize=tiny]{todonotes}

\icmltitlerunning{\ourtitle: An Active Defense Solution for DNN Model via Automated Weight Obfuscation}

\begin{document}

\twocolumn[
\icmltitle{\ourtitle: An Active Defense Solution for DNN Model via Automated Weight Obfuscation}




\begin{icmlauthorlist}
\icmlauthor{Tong Zhou}{neu}
\icmlauthor{Yukui Luo}{neu}
\icmlauthor{Shaolei Ren}{ucr}
\icmlauthor{Xiaolin Xu}{neu}

\end{icmlauthorlist}

\icmlaffiliation{neu}{Northeastern University, Boston, MA}
\icmlaffiliation{ucr}{UC Riverside, Riverside, CA}

\icmlcorrespondingauthor{Xiaolin Xu}{x.xu@northeastern.edu}

\icmlkeywords{Deep Neural Network, Model Protection}

\vskip 0.3in
]



\printAffiliationsAndNotice{}  

\begin{abstract}
As a type of valuable intellectual property (IP), deep neural network (DNN) models have been protected by techniques like watermarking. However, such passive model protection cannot fully prevent model abuse. 
In this work, we propose an active model IP protection scheme, namely \ourtitle, which actively protects the model by splitting it into two parts: the \textit{obfuscated model} that performs poorly due to weight obfuscation, and the \textit{model secrets} consisting of the indexes and original values of the obfuscated weights, which can only be accessed by authorized users with the support of the trusted execution environment.
Experimental results demonstrate the effectiveness of \ourtitle, e.g., by only modifying 275 out of over 11 million (i.e., 0.002\%) weights, the accuracy of the obfuscated ResNet-18 model on CIFAR-10 can drop to 10\%. 
Moreover, \ourtitle is stealthy and resilient against norm clipping and fine-tuning attacks, making it an appealing solution for DNN model protection.
The code is available at: \url{https://github.com/Tongzhou0101/NNSplitter}.
\end{abstract}

\section{Introduction}
\label{sec:intro}

Despite the success of deep neural networks (DNNs) in various applications \cite{duong2019mobiface,wang2018pelee}, building a DNN model with high accuracy is costly, i.e., requiring a large number of labeled samples and massive computational resources \cite{jiang2020unified}. As a result, a high-performance DNN model presents valuable intellectual property (IP) of the model owner, which should naturally be adequately protected against potential attacks. However, recent studies have demonstrated that millions of on-device ML models are vulnerable to model IP attacks \cite{sun2021mind}, wherein the attacker can extract the model and deploy it on unauthorized devices. Such unauthorized usage leads to significant financial losses for the model owners.

Several studies have addressed the issue of DNN model protection, which can be broadly classified into two categories:  passive protection (after IP infringement) and active protection (before IP infringement). 
Although passive protection techniques, e.g., watermarking, help model owners declare the ownership and guard their rights \cite{yang2021robust,zhang2018protecting}, they cannot effectively prevent unauthorized usage as the model can perform very well in most cases. Thus, attackers are still motivated to steal the well-performed model and use it without  the knowledge  of the model owner.

\begin{table}[t]
    \centering
    \resizebox{0.95\linewidth}{!}{
    \begin{tabular}{ll}
    \toprule
     \textbf{Requirements}    &  \multicolumn{1}{c}{Definitions}\\ \midrule
     \textbf{Effectiveness}    & \begin{tabular}[c]{@{}l@{}}The obfuscated model exhibits poor per-\\formance (e.g., random-guess accuracy).\end{tabular}\\
     \midrule
     \textbf{Efficiency}    &  \begin{tabular}[c]{@{}l@{}}The number of model secrets stored in \\ the secure space should be minimized.\end{tabular}\\
     \midrule
     \textbf{Integrity}    & \begin{tabular}[c]{@{}l@{}}The functionality of the model is  \\ preserved for the legitimate users. \end{tabular}\\
     \midrule
     \textbf{Resilience}  & \begin{tabular}[c]{@{}l@{}l@{}}The obfuscated model should be \\ resilient against potential attack surfaces. \end{tabular}\\
     \midrule
     \textbf{Stealthiness} &\begin{tabular}[c]{@{}l@{}} The obfuscated weights should be \\ indistinguishable from normal weights.\end{tabular}\\
    \bottomrule
    \end{tabular}}
    \caption{The design requirements for an efficient model protection scheme, and the guidance for our proposed \ourtitle.}
    \label{tab:requirments}
\end{table}

In contrast, active protection only allows legitimate users to use the well-performed model, while intentionally degrading the model functionality for attackers, thus protecting the interests of the model owner \cite{chakraborty2020hardware, fan2019rethinking, zhou2022obfunas}. Nonetheless, such an advantage of the active protection methods is not free, which either requires hardware support, e.g., a hardware root-of-trust \cite{chakraborty2020hardware}, or introduces extra model parameters \cite{fan2019rethinking}. 
Moreover, the existing active protection approaches are not generic, i.e., they require special training strategies for model protection, rendering them inapplicable to pre-trained models.  It is also worth noting that some fault injection methods can also cause accuracy deterioration \cite{liu2017fault}, using software-oriented \cite{rakin2019bit} or hardware-oriented \cite{luo2021deepstrike,rakin2021deep} attacking schemes. However, the design of those works is from the perspective of an attacker, which can not satisfy requirements (shown in Tab.~\ref{tab:requirments}) for active protection,  with detailed discussion in Sec.~\ref{sec:performance}.

Considering these limitations of existing defense strategies, we are motivated to develop a generic active model IP protection scheme. Specifically, we propose to split the victim model into an obfuscated model and model secrets, which should fulfill the requirements detailed in Tab.~\ref{tab:requirments}. 
The design of such a scheme presents the following substantial challenges (\textbf{C}). \textbf{C1:} Given the limited size of secure memory we can leverage, e.g., the trusted execution environment (TEE) \cite{costan2016intel}, the stored model secrets need to be kept small, while there are millions of, if not more, weights in modern DNN models.
\textbf{C2:} The model functionality should be preserved for legitimate users. 
\textbf{C3:} The obfuscated weights should be imperceptible and not easily identified by attackers.
\textbf{C4:} Attackers can not significantly improve the degraded accuracy with reasonable efforts.

To address \textbf{C1}, our proposed scheme, namely \textit{\ouralg}, generates a mask that selectively obfuscates weights within a small range. This range is chosen to be small enough, so that the original values of the obfuscated weights can be replaced by a single value, thereby reducing the storage requirements for model secrets. To achieve this goal, we utilize a reinforcement learning (RL) algorithm to design a controller that efficiently identifies important filters with significant influences on model predictions. By focusing on these filters, we can minimize the number of obfuscated weights while still achieving a significant accuracy drop. 
For \textbf{C2}, we profile the model weights and adjust the aforementioned small range to ensure that the original model accuracy can be preserved, after applying the obfuscated weights restoration rule (details are given in Sec. \ref{sec:problem}). Besides, we set a limit to ensure the obfuscated weights remain within the original weight range to avoid being identified by attackers (addressing \textbf{C3}). Last, we force the weight changes to spread across various layers to increase the resilience against potential attack surfaces to improve accuracy (addressing \textbf{C4}).

Overall,  \ouralg achieves model IP protection by splitting a victim model into two parts: the \textit{obfuscated model} and the \textit{model secrets}. Specifically, the obfuscated model is vulnerable to model extraction, but its degraded accuracy resulting from weight obfuscation renders it practically useless, effectively mitigating the vulnerability.  Meanwhile, the model secrets are secured by TEE to provide authorized inference, which can only be accessed by authorized users.  
The contributions of this work are as follows:
\begin{itemize} 
\setlength\itemsep{-0.2em}
    \item We systematically define the requirements for active model protection and propose \ouralg that can automatically split the victim model into the obfuscated model and model secrets with all of these design requirements fulfilled.    
    \item The accuracy of the obfuscated model can drop to random guess by  modifying only 0.001\% weights ($\sim$ 300) of the victim model, which is hardware-friendly due to low secure memory requirement.    
    \item We demonstrate that the proposed \ouralg is resilient against potential attacks, including norm clipping and fine-tuning attacks.
\end{itemize}

\section{Related Works and Background} 
\label{sec:background}
\subsection{Threat Model}\label{sec:threat_model}
To ensure highly effective model protection, we  consider a strong attacker who has the capability to extract the exact victim DNN model, including its architecture and model parameters, using techniques like in-memory extraction mentioned in \cite{sun2021mind}. For example,  attackers can download a mobile application built with a DNN model, de-compile it, extract the model file, and deploy it on their own devices. Besides, we assume the attackers only have limited training data; otherwise, they can train a competitive model on their own, without strong incentives to steal the victim model. By considering these scenarios, we aim to design a model protection scheme that can effectively safeguard the victim model IP against such strong attackers.

\subsection{Trusted Execution Environment}
While passive model IP protection fails to protect models from being stolen or used, we envision the TEE (e.g., ARM TrustZone on mobile devices \cite{ngabonziza2016trustzone}) as a promising solution to achieve active model protection. TEE provides a physical isolation scheme in the hardware devices that separates memory into the normal (untrusted) world and the secure (trusted) world, where the normal world can communicate with the secure world by invoking a secure monitor call \cite{ye2018tzslicer}. 
This setup ensures that only legitimate users can access the secure world, while attackers are blocked. 
Given the effectiveness of TEE in model protection as demonstrated in previous works  \cite{chen2019deepattest, shadownet}, we adopt the TEE implementation scheme following \cite{shadownet} without delving into the technical details or considering the vulnerability of TEE (e.g., side-channel attacks), as it is not the primary focus in this work.

It is important to note that the secure memory of TEE is limited, e.g., $\sim$ 10 MB for trusted applications \cite{shadownet}. On the other hand, the size of state-of-the-art (SOTA) DNN models continues increasing, e.g., large models like ResNet-101 exceed 155M parameters \cite{he2016deep}. To accommodate this limitation, our approach \ouralg aims to obfuscate as few weights as possible, to minimize the overhead on secure memory usage.

\subsection{Model IP Protection}
The existing literature has actively addressed model security issues on edge devices \cite{sun2021mind, xu2019first,shukla2021device}, 
and demonstrated that attackers can easily extract the model even without sophisticated skills \cite{sun2021mind}.
As discussed above, the existing passive model protection methods like watermarking \cite{yang2021robust} have limitations in fully preventing model piracy. 
On the other hand, active protection methods, such as model encryption \cite{al2020survey}, have been proposed where the model files are encrypted and stored in memory. However, the encrypted model needs to be decoded at runtime for inference, which can still be vulnerable to attacks. 

To enhance the model IP security, Chakraborty \etal leverage secure hardware support and propose a key-dependent back-propagation algorithm to train a DNN architecture with the weight space obfuscated \cite{chakraborty2020hardware}. After obfuscation, only authorized users are allowed to use the model on trusted hardware with the key embedded on-chip, while the model accuracy will drop significantly if attackers extract the model and deploy it on other devices. However, this method requires hardware modification and cannot be generally used to protect pre-trained models. 
Similarly, Fan \etal~propose a method to protect the model IP by embedding a passport layer within the DNN model, so that the DNN inference performance of an original task will be significantly deteriorated due to forged passports \cite{fan2019rethinking}. However, this work aims to defend against ambiguity attacks, and can only be applied to the models already embedded with watermarks.
These existing approaches provide valuable insights into model protection, but they either require hardware modifications or have specific limitations in their applicability. 

\subsection{Difference from Fault Injection}
\label{sec:performance}

A key point of active model protection is to introduce performance degradation (e.g., accuracy drop) into the protected model. 
Although the objective is similar to fault injection attacks that manipulate the DNN model parameters to cause abnormal inference \cite{liu2017fault}, the fundamental design requirements are largely different: (i) \textbf{Stealthiness:} fault injection attacks do not consider stealthiness in model manipulations, which introduce extremely large magnitudes changes and can be easily distinguished and removed by applying weights range restriction \cite{chen2021low,liu2017fault}. (ii) \textbf{Resilience:} most fault injection attacks only target the most direct parameters of outputs, e.g., those in the last layer. However, such an attack is not resilient against fine-tuning. 
Also, although existing attacks like bit-flip \cite{rakin2019bit} modify the weight bits in different layers to degrade model accuracy, such gradient-ranking-based attacks can be mitigated by weights reconstruction \cite{li2020defending}. Moreover, bit-flip targets the quantized DNN models, where the weight magnitude is constrained based on the quantization method, 
while how to ensure the stealthiness and resilience of attacks on the floating-point precision DNN models is significantly under-explored.

In sharp contrast to these studies on attacks, we 
rethink and address all the aforementioned design limitations from a defense perspective. 
Specifically, we mainly explore an active defense scheme leveraging hardware support from the TEE, to actively prevent attackers from obtaining functional DNN models and make such model extraction attacks less motivated. Our work is orthogonal to the existing literature and can be generally applied to any pre-trained models.

\begin{figure*}[t]
    \centering    \includegraphics[width=0.98\textwidth]{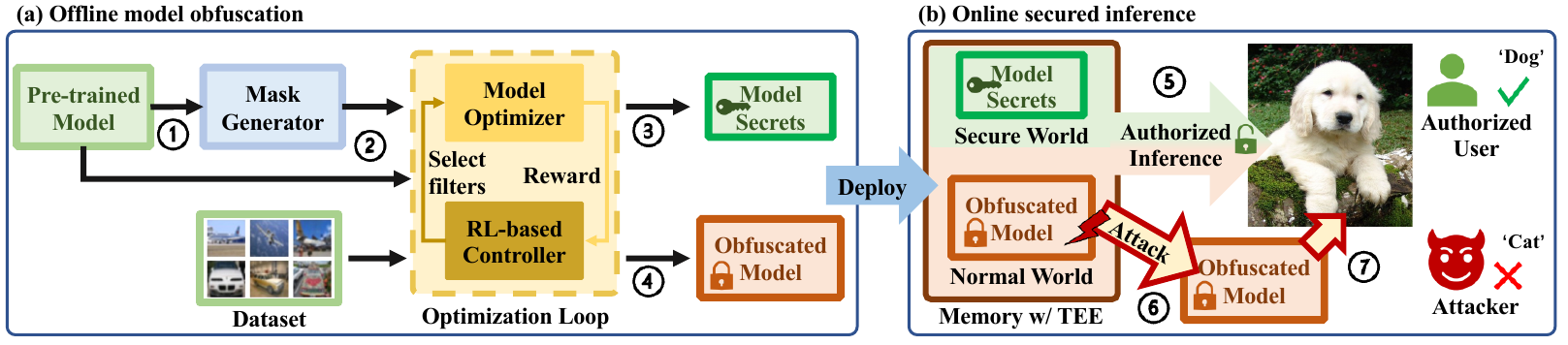}
    \caption{An overview of \ourtitle. (a) Offline model obfuscation: \ouralg splits the pre-trained model into two parts once the reward is converged, i.e., the obfuscated model and the model secrets (including the indexes and the original weight values).  
    (b) Online secured inference: an attacker can only extract the obfuscated model  stored in the normal world, which exhibits poor performance. However, the original model accuracy can be preserved by integrating the model secrets stored in the secure world of the victim's device.}
    \label{fig:overview}
\end{figure*}

\section{Our Proposed Method: \ourtitle}
\label{sec:method}
This section presents our proposed active DNN model protection method, \ourtitle,  
which meets the requirements of effectiveness, efficiency, integrity, resilience, and stealthiness, as described in Tab.~\ref{tab:requirments}.

The overview of \ourtitle is illustrated in Fig. \ref{fig:overview}, including the offline model obfuscation and the online secured inference. In the offline phase, taking the pre-trained DNN model as input (\textcircled{\small{1}}), the mask generator profiles the weight distribution to determine the parameters of the mask following certain rules (Sec. \ref{sec:problem}). The mask parameters and the DNN model will be fed into the optimization loop (\textcircled{\small{2}}) along with the dataset. In the loop, we build a RL-based controller to help form a filter-wise mask, which is used to guide the weight obfuscation optimized by the model optimizer. 
Then the negative accuracy evaluated on the test dataset will serve as a reward to optimize the controller.  When the reward converges, i.e., the accuracy stops decreasing, the optimization loop will generate two parts --- the obfuscated model (\textcircled{\small{4}}) deployed in the normal world (untrusted memory), and the  model secrets (\textcircled{\small{3}}) that include the
indexes and the original values of the obfuscated weights stored in the secure world (trusted memory).

During online secured inference, the model is executed layer by layer.  At each layer, the obfuscated weights are used to compute an output feature map, which may contain errors in certain output channels. These errors are intentionally propagated to subsequent layers, resulting in a substantial drop in accuracy. 
This mechanism effectively prevents unauthorized use by attackers, as the model they extract from the normal world will perform poorly due to the presence of obfuscated weights (\textcircled{\small{6}} and \textcircled{\small{7}}).
However, in the secure world, the model secrets are utilized to correct these errors in the specific output channels, ensuring that the model functions as intended for legitimate users who have access to the secure world (\textcircled{\small{5}}). 
In this way, the majority of the DNN inference computation is performed in the main memory of the normal world, reducing the computation overhead within the secure world.

\subsection{Problem Formulation}
\label{sec:problem}
Given a pre-trained DNN model $\mathcal{M}$ containing $L$ convolutional/fully connected layers with weights $\textbf{W}:= \{\textbf{W} ^{(l)}\}_{l=1}^L$, 
we aim to find the optimal \textit{weight changes} $\Delta \textbf{W}$ (the same size as $\textbf{W}$) that maximize the classification loss function $\mathcal{L}_\mathcal{M}$. For simplicity, we denote each element in $\textbf{W}$ and $\Delta \textbf{W}$ as $w_{i}$ and $\Delta w_{i}$, respectively, where $i\in [1, N]$ and $N$ is the total number of model weights. 
Upon achieving the optimal weight obfuscation, we store the indexes of non-zero $\Delta w_{i}$ and original $w_i$ to preserve the performance of the victim model for legitimate users.

\textbf{Mask Generator.} To reduce the secure storage requirement, we design a mask $\textbf{M}$ for $\Delta \textbf{W}$ to determine the weights to be obfuscated, which is defined by:
\begin{equation}
\label{eq:mask}
  \textbf{M}(w_i) = 
  \begin{cases}
  1 & if \left|w_{i}-c\right| \le \epsilon, \\
  0 & otherwise,
  \end{cases}
\end{equation}
where $c$  and $\epsilon$ are both controllable hyper-parameters. Using this mask, we can refine the weight changes $\Delta$\textbf{W}$'$ := $\Delta$\textbf{W} $\odot$ \textbf{M}, where $\odot$ denotes element-wise multiplication. The benefits of the mask design are two-fold: \textbf{(i)} $\textbf{M}$ only allows weights in the range [$c-\epsilon, c+\epsilon$] to be obfuscated. By selecting a small $\epsilon$, we ensure that the obfuscated weights are close to a constant value $c$. This allows us to store a single value for these obfuscated weights instead of multiple different values, thus saving the secure space while preserving the model functionality; \textbf{(ii)} by carefully selecting $c$, we can distribute  the weight obfuscation across various layers, significantly improving the resilience against the potential attack surfaces, such as fine-tuning (see results in Sec. \ref{sec:attack}). 
Besides, we apply $\ell_0$-norm regularization to $\Delta \textbf{W}$$'$ to further save the secure storage space.

\textbf{Model Optimizer.} To improve the stealthiness of weight obfuscation, we restrict the values of \textit{obfuscated weights}, i.e., $\textbf{W}+\Delta \textbf{W}'$, within the original value range of $\textbf{W}$, which is achieved by the hyperparameters $\alpha$ and $\beta$ in Eq.~(\ref{eq:objective}). Thus, the optimal $\Delta \textbf{W}'$ can be found by minimizing the loss function $\mathcal{L}(\Delta \textbf{W}')$:

\vspace{-0.13in}
{\small
\begin{equation}
\label{eq:objective}
\begin{split}
   \min _{\Delta \textbf{W$'$}} \mathcal{L}(\Delta \textbf{W$'$}) &=-\mathcal{L}_\mathcal{M}\left(f\left(\boldsymbol{x} ;\textbf{W}+\Delta \textbf{W$'$}  \right), \boldsymbol{y}\right) +
 \lambda \left\| \Delta \textbf{W$'$}\right\|_0\\
 \text { s.t. } \enspace \alpha * &\min\{w_{i}\} \leq w_{i} + \Delta w'_{i} \leq \beta * \max\{w_{i}\} \enspace \forall i,
\end{split}
\end{equation}
}
where $f$ denotes the functionality
of the DNN model $\mathcal{M}$, $\boldsymbol{x}$ is the training samples with $\boldsymbol{y}$ being the corresponding labels, and $\lambda$ controls the sparsity of weight changes.

However, considering the SOTA DNN models consisting of millions of parameters, only using $\ell_0$-norm to minimize the number of weight changes is not sufficient. Inspired by the fact that the importance of different filters varies \cite{you2019gate}, e.g., the filters learning the background features contribute less compared to these learning the object edge, we propose to embed the filter-wise weights selection strategy into the mask design. This strategy involves adding weight changes only to selected important filters while still satisfying the constraints in Eq. (\ref{eq:mask}). By doing so, we can further reduce the storage space required for weight obfuscation, while still achieving the desired level of accuracy degradation.

Nonetheless, manually selecting filters to design an optimal filter-wise mask is impractical due to the large number of filters in SOTA DNNs. Thus, we propose a RL-based controller to automatically select the optimal filters. 

\subsection{RL-based Controller}
As an important component of \ourtitle, the RL-based controller aims to form  a filter-wise mask.
While a straightforward approach would be to use the controller to generate all the hyperparameters required by the design of $\textbf{M}$, including $c$ and $\epsilon$ in Eq.~(\ref{eq:mask}), this design principle would increase the complexity and optimization difficulty of developing the controller.
To overcome the challenges while maintaining the effectiveness of the controller, we leverage the domain knowledge about the distribution of the model weights to determine the values of these two hyper-parameters (see details in Sec.~\ref{sec:para_set}), and leave the difficult part, i.e., selecting important filters,  to the controller.

The developed controller consists of three parts: an encoder for encoding the initialized state, a policy network for decision-making, and decoders for different layers to decode the output of policy networks into filter indexes. In this controller, an agent  selects a filter with index $k$ for each layer (i.e., actions), where $k\in [1,  K^{(l)}]$ and $K^{(l)}$ denotes the number of filters (i.e., output channels) of the $l$-th layer. Since the state $K^{(l)}$ is determined by the architecture of the victim model $\mathcal{M}$, the environment is static for the agents.
To select $n$ filters for each layer ($n$ could be 1), we will have $n$ agents making $n*L$ actions in total, denoted as $a_{1:n*L}$. All agents will share the same controller with weights $\theta$, which will be optimized by maximizing the expected reward $J(\theta)$:
\begin{equation}
    J(\theta) = E_{\pi(a_{1:n*L};\theta_c)}[R], 
    \label{eq:exp_reward}
\end{equation}
where $\pi(\cdot)$ denotes the probabilities of taken actions given $\theta$, and reward $R$ is constructed by the negative inference accuracy of the obfuscated model, 
defined by Eq. (\ref{eq:reward}):
\begin{equation}
    R = -ACC\left(f\left(\boldsymbol{x^{t}} ;\textbf{W}+\Delta \textbf{W$'$}),\boldsymbol{y^{t}}\right)\right),
    \label{eq:reward}
\end{equation}
where $ACC$ is accuracy, $\boldsymbol{x^{t}}$ is the validation dataset and $\boldsymbol{y^{t}}$ denotes the corresponding labels. Considering $R$ is non-differentiable with respect
to the controller output, we use a policy gradient method: REINFORCE algorithm \cite{williams1992simple} to maximize $J(\theta)$, which is the same as minimizing the loss function of the controller: 
\begin{equation}
    \mathcal{L}_c(\theta) = - \frac{1}{m} \sum_{j=1}^{m} \sum_{t=1}^{n*L} \log \pi\left(a_{t}; \theta_{c}\right)\left(R_{j}-b\right),
    \label{eq:controller}
\end{equation}
where $m$ represents the number of trails in each episode of the controller, and $b$ denotes an exponential moving average of the rewards used to reduce the variance for updating $\theta$. 

The obfuscated model generation is described in Alg. \ref{alg:overall}. With the mask parameters $c$ and $\epsilon$ obtained from the mask generator (line \ref{alg:mask}), the initialized controller will first design a filter-wise mask to optimize the victim model by minimizing the Eq.~(\ref{eq:objective}) (line \ref{alg:act}-\ref{alg:opt}), then the controller use rewards obtained from the victim model to optimize itself (line \ref{alg:reward}-\ref{alg:con}). When the reward converges,  \ouralg will output two parts, which are the obfuscated model and model secrets that will be stored in the secure world.

\begin{algorithm}[t]
\caption{Offline obfuscated model generation}
\label{alg:overall}
\textbf{Input}: pre-trained model $\mathcal{M}$ with weights $\mathbf{W}$; initialized controller with $\theta$; training data ($\boldsymbol{x},\boldsymbol{y}$); test data ($\boldsymbol{x^{t}},\boldsymbol{y^{t}}$); $K^{(l)}$, $\alpha$, $\beta$, $\lambda$.\\
\textbf{Parameters}: learning rate $\eta_1$, $\eta_2$.\\
\textbf{Output}: model secrets (the indexes of $\Delta w'_i$ and c),  obfuscated model $\mathcal{M'}$. 
\begin{algorithmic}[1] 
\State Feed $\mathcal{M}$ into model generator and obtain $c$ and $\epsilon$ \label{alg:mask}
\Repeat
\Statex \textcolor{gray}{// Optimization loop}
\For{m batches}
\State Use controller to generate filter indexes \label{alg:act}
\State Form filter-wise mask $\mathbf{M}$ and feed into model optimizer 
\Statex \textcolor{gray}{// Optimize $\mathbf{\Delta W'}$}
\For {training epochs}
\State Minimize $\mathcal{L}(\Delta \textbf{W$'$})$ 
\Comment{\textcolor{Green}{Eq. (\ref{eq:objective})}}
\State Update $\Delta \textbf{W$'$} \leftarrow \Delta \textbf{W$'$} -  \eta_1 \nabla \mathcal{L}(\Delta \textbf{W$'$})$ \label{alg:opt}
\State Measure accuracy on ($\boldsymbol{x^{t}},\boldsymbol{y^{t}}$) \label{alg:reward}
\State Collect the reward $R$ \Comment{\textcolor{Green}{Eq. (\ref{eq:reward})}}
\EndFor 
\Statex \textcolor{gray}{// Optimize the controller $\theta$}
\State Calculate the average reward b
\State Minimize $\mathcal{L}_c(\theta)$ 
\Comment{\textcolor{Green}{Eq. (\ref{eq:controller})}}
\State Update controller: $\theta \leftarrow \theta - \eta_2 \nabla \mathcal{L}_c(\theta)$ \label{alg:con}
\EndFor
\Until{Reward R is converged}
\end{algorithmic}
\end{algorithm}

\section{Experimental Validation}

\label{sec:experiments}

\subsection{Experimental Setup}
\label{sec:setup}

\textbf{Datasets.} We evaluate the effectiveness of \ourtitle on models trained with three datasets: Fashion-MNIST \cite{xiao2017fashion}, CIFAR-10, and CIFAR-100 \cite{cifar10_data}. For Fashion-MNIST, there are 60k 28 $\times$ 28 grayscale images from 10 classes in the training dataset and 10k images in the test dataset. Besides, CIFAR-10/100 both have 50k training images and 10k test images of 32 $\times$ 32, except that CIFAR-10 includes 10 classes while CIFAR-100 has 100 classes.

\textbf{Baseline DNN Models.} While \ouralg applies to any pre-trained models, here we consider several commonly-used DNNs as the proof-of-concept, including VGG-11 \cite{simonyan2014very}, MobileNet-v2 \cite{sandler2018mobilenetv2}, and ResNet-18/20 \cite{he2016deep} trained on the aforementioned datasets. To demonstrate that \ouralg is a generic defense solution regardless of the victim model's training strategies, i.e., training-free, we use pre-trained models with weights public online, where the parameter settings (e.g., layer dimensions) could be different for the same DNN class for different datasets. We use the structures and pre-trained weights as they are released online, \textit{despite that they may not reach the best-known accuracy on these datasets}.

\textbf{Comparison Methods.} 
Since there are no existing works that follow the same settings and objectives as \ouralg, we propose the following methods for comparison in order to demonstrate its effectiveness. (i) \textbf{Random:} instead of using domain knowledge and the RL-based controller to design a filter-wise mask, we assume a model protection method that randomly generates a binary mask to select weights and obfuscate them by optimizing Eq. (\ref{eq:objective}). For a fair comparison, the binary mask will select the same number of obfuscated weights as \ouralg. (ii) \textbf{Base-\ouralg}: this method randomly selects filters in each layer instead of using the RL-based controller to optimize the selection.

\subsection{Hyper-parameters Setting}
\label{sec:para_set}

\textbf{Weight Constraints.} To enhance the stealthiness of weight changes, we use two hyper-parameters $\alpha$ and $\beta$ in Eq. (\ref{eq:objective}) to ensure that the values of \textit{obfuscated weights} are indistinguishable from the normal weights, thereby  avoiding outlier detection. Considering $\min\{w_{i}\} < 0$ and $\max\{w_{i}\} > 0$ in general, the values of $\alpha$ and $\beta$ are in the range (0,1].  Specifically, they are set to 0.95 for the following experiments.

\begin{figure}[htbp]
    \centering
    \includegraphics[width=0.45\textwidth]{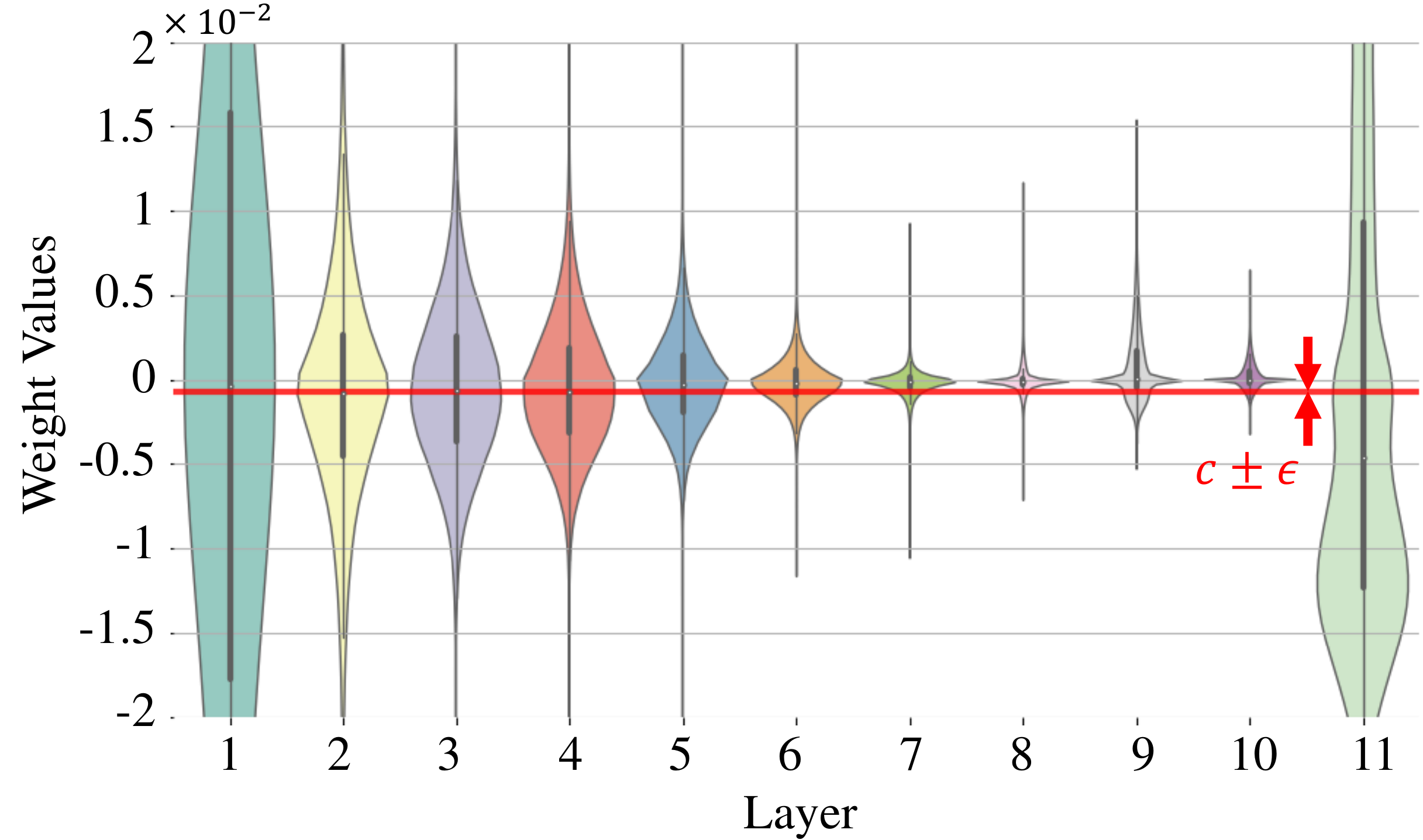}
    \caption{The obfuscated weights spread across all layers, illustrated in a VGG-11 model trained on CIFAR-10 as an example.}
    \label{fig:layer_dis}
\end{figure}

\textbf{Mask Design.} The mask design depends on the domain knowledge of the weight distribution. Specifically, to determine the mask hyper-parameters $c$ and $\epsilon$, we profile the weight distribution of each layer and take the average of the median values as the $c$, which will encourage the weight changes to spread across various layers, as shown in Fig. \ref{fig:layer_dis}. As for determining $\epsilon$, the principle is to ensure that the accuracy can be preserved when replacing weights in the range of [$c-\epsilon, c+\epsilon$] with $c$. Therefore, the closer $c$ is to the median of the total weights, the smaller $\epsilon$ should be. Otherwise, the baseline accuracy cannot be restored due to a great precision loss. The details are shown in Tab. \ref{tab:mask para}.

\begin{table}[htbp]
\centering
\renewcommand\tabcolsep{13pt}
\resizebox{0.85\linewidth}{!}{
\begin{tabular}{cccc}
\toprule
\multirow{2}{*}{Dataset} & \multirow{2}{*}{Model} & \multicolumn{2}{c}{Hyper-parameters}                                  \\
\cmidrule(r){3-4} 
                 & &  \multicolumn{1}{c}{$c$} & \multicolumn{1}{c}{$\epsilon$} 
                 \\ \midrule
\multirow{3}{*}{\begin{tabular}[c]{@{}c@{}}Fashion \\ MNIST\end{tabular}}  
 & VGG-11           &   -1.7e-3     &         1e-4                              \\ 
& ResNet-18         &                             -2.3e-4  &  3e-5                         \\ 
& MobileNet-v2       &                        -3.8e-4      &         4e-3 
                 \\ \midrule
\multirow{3}{*}{CIFAR-10} 
 & VGG-11                &          -7.0e-4                    &    1e-4        \\ 
& ResNet-18              &          -1.2e-3                     &    8e-5                       \\ 
& MobileNet-v2      &         -2.5e-4                     &    1e-4                          \\ \midrule
\multirow{3}{*}{CIFAR-100} 
& VGG-11           &                 -1.9e-3         &    5e-4        \\ 
& ResNet-20        &         -6.7e-3                       &    6e-4                       \\ 
& MobileNet-v2     &       -9.4e-4                     &   3e-4  \\          
\bottomrule
\end{tabular}}
\caption{The settings of mask hyper-parameters. }
\label{tab:mask para}
\end{table}

\textbf{Controller Design.} The RL-based controller in our approach follows a similar design to
the neural architecture search in \cite{zoph2016neural, zoph2018learning, pham2018efficient}, i.e., using a recurrent neural network (RNN) to build the policy network, where the embedding dimension and the hidden dimension of the RNN policy network are set to 256 and 512, respectively. Besides, we use one-hot encoding to encode the initialized state as the input of the policy network. 
For decoding the output of the policy network into filter indexes, we build a decoder for each layer in the DNN victim model with a linear layer, where its output dimension is equal to the number of output channels in the corresponding DNN layer.

\begin{table*}[!t]
\centering
\resizebox{0.85\linewidth}{!}{
\begin{tabular}{cccccccc}
\toprule
\multirow{2}{*}{Dataset} & \multirow{2}{*}{Model} & \multicolumn{2}{c}{Baseline}                                  & \multicolumn{1}{c}{Obfu. Weights}    & \multicolumn{2}{c}{Obfu. Acc. (\%)}                                        
 & \multicolumn{1}{c}{\multirow{2}{*}{\begin{tabular}[c]{@{}c@{}}Restored \\ Acc. (\%)\end{tabular}}}
 \\
\cmidrule(r){3-4}    \cmidrule(r){6-7}  
                  & &\multicolumn{1}{c}{Acc. \tablefootnote{Acc. denotes the top-1 accuracy for all cases.} (\%)} & \multicolumn{1}{c}{Para. (M)} & 
                  Num. / Ratio \tablefootnote{Ratio is calculated by the number of obfuscated weights divided by the total number of model parameters.}(\%)
                   &\multicolumn{1}{c}{\textbf{\ourtitle}} &  \multicolumn{1}{c}{Random}                         & \multicolumn{1}{c}{}                                     
                 \\ \midrule
\multirow{3}{*}{\begin{tabular}[c]{@{}c@{}}Fashion \\ MNIST\end{tabular}}  
 & VGG-11            &   93.73    &  28.14     &       \textbf{313/ 0.001}             &   \textbf{10.00}    &   92.90$\pm$0.40     &    93.73                                    \\ 
& ResNet-18         &  93.71    &  11.17     &       \textbf{231/ 0.002}              &     \textbf{10.00}      &    92.03$\pm$0.93    &         93.71                         \\ 
& MobileNet-v2     &    93.97    &  2.24      &      \textbf{340/ 0.015}               &    \textbf{10.00}       &    86.41$\pm$1.58    &    93.97               
                 \\ \midrule
\multirow{3}{*}{CIFAR-10} 
 & VGG-11            &   92.39    & 28.15     &       \textbf{876/ 0.003}             &           \textbf{10.78}    &    91.42$\pm$0.25    &    92.39           \\ 
& ResNet-18         &  93.07    & 11.17     &       \textbf{275/ 0.002}              &      \textbf{10.00}      &  91.35$\pm$0.27      &         93.07                 \\ 
& MobileNet-v2     &     93.91    & 2.24      &      \textbf{835/ 0.037}               &       \textbf{10.48}    &     78.18$\pm$1.38   &         93.91                      \\ \midrule
\multirow{3}{*}{CIFAR-100} 
& VGG-11            &   70.50   &   9.80  &      \textbf{782/ 0.008}              &   \textbf{1.34}    &   64.34$\pm$0.75     &    70.50              \\ 
& ResNet-20        &   68.28    & 0.28     &     \textbf{96/ 0.034}               &      \textbf{1.31}    &     56.35$\pm$1.38   &         68.27         \\ 
& MobileNet-v2     &     74.29    & 2.25      &   \textbf{447/ 0.019}                   &      \textbf{1.00}    &   50.92$\pm$1.33     &   74.28        \\          
\bottomrule
\end{tabular}}
\caption{\ourtitle applied to multiple  DNN models on three datasets. The number of obfuscated weights is the median value when the obfuscated (Obfu.) accuracy degraded to random guess ($<$11\% for Fashion-MNIST/CIFAR-10, and $<$2\% for CIFAR-100). The obfuscated accuracy of random is reported as mean$\pm$std with the same number of obfuscated weights.}
\label{tab:result}
\end{table*}

\subsection{Performance Evaluation}
To find the optimal changes added to the pre-trained models, we leverage the designed RL-based controller to select filters in both convolutional layers and fully connected layers. Here, we also refer to each output channel of the fully connected layer as a filter for simplicity. 
The results of \ouralg, baseline, and \textit{random} methods are shown in Tab. \ref{tab:result}. Following our defined requirements for DNN model protection schemes in Tab. \ref{tab:requirments}, we evaluate the performance of \ourtitle from three perspectives: effectiveness, efficiency, and integrity.

\textbf{Effectiveness.} As shown in column 5 and column 6 in Tab. \ref{tab:result}, \ouralg successfully degrades the victim model inference accuracy to random guessing, rendering the attacker's effort useless.  Specifically, for 10-class datasets like Fashion-MNIST and CIFAR-10, the obfuscated top-1 accuracy of all victim models is lower than 11\%, while for CIFAR-100 including 100 classes, the top-1 accuracy of victim models after obfuscation is lower than 2\%.
In contrast, randomly selecting weights to achieve model obfuscation only causes a limited accuracy drop (column 7 in Tab. \ref{tab:result}), e.g., $\sim$1\% (92.90$\pm$0.40 \% vs. 93.73\%) accuracy drop for VGG-11 model trained on Fashion-MNIST.
Besides, the number of obfuscated weights is below 1k for all cases, which is small enough to store in TEE \cite{costan2016intel}. The smaller storage requirement can support more models deployed on the same device.

\textbf{Efficiency.} Given the ever-increasing size of DNN models, we aim to achieve active model protection by modifying only a very small fraction of the model weights. 
Specifically, by obfuscating 0.001\% weights of the VGG-11 model on Fashion-MNIST, the model becomes completely malfunctional, i.e., with inference accuracy equal to random guess. Besides, for even more complicated datasets like CIFAR-100, the ratio of weight obfuscation is still small, e.g., 0.008\% for VGG-11. Note that our proposed design could further reduce this ratio by tuning the mask hyper-parameters $c$ and $\epsilon$. However, for a \textit{fair comparison}, we follow the generic strategy for all victim models to determine these parameters as described in Sec. \ref{sec:para_set}.

\begin{figure}[hbpt]
    \centering
    \includegraphics[width=0.44
    \textwidth]{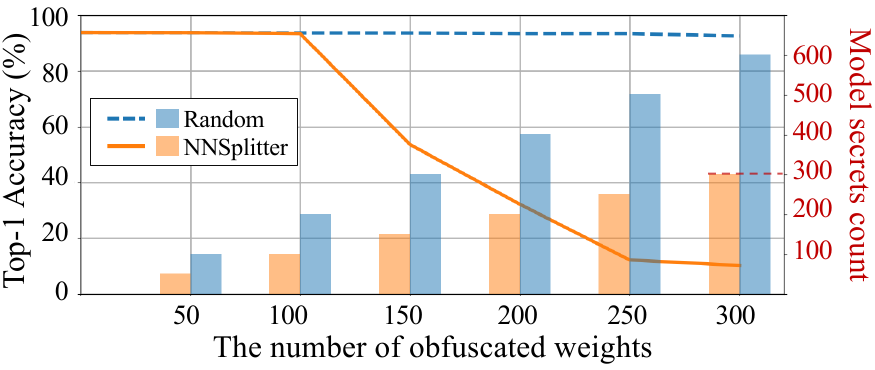}
    \caption{Number of obfuscated weights vs. accuracy for VGG-11 trained on CIFAR-10 (line plot), with the number of corresponding model secrets shown in the bar plot (associated with the  y-axis on the right).}
    \label{fig:cnt_acc_cf10}
\end{figure}

Furthermore, Fig. \ref{fig:cnt_acc_cf10} demonstrates that fewer weight changes are required when the desired accuracy degradation is smaller. For example, with 300 obfuscated weights and 301 model secrets (including 300 indexes and the value of $c$), \ouralg achieves an accuracy drop to 10.23\% for the VGG-11 model on Fashion-MNIST. Besides, a noticeable accuracy drop can still be observed as the number of obfuscated weights decreases to 150. In contrast, by randomly obfuscating 300 weights, the accuracy only drops to 92.49\%, while the number of secrets is almost doubled, i.e., 600, since the original values of obfuscated weights are not close and thus cannot be replaced by a single value.

\textbf{Integrity.} Ensuring normal model inference for legitimate users is essential for an active model protection method. Thus our method should securely eliminate the adverse effects of obfuscated weights for authorized use. Specifically, with access to the model secrets stored in TEE, the obfuscated weights can be located according to stored indexes. 
Since our proposed method carefully selects weights within [$c-\epsilon, c+\epsilon$] with a very small $\epsilon$ (reported in Tab. \ref{tab:mask para}), we can replace the constant $c$ with the obfuscated weights during computation, thus preserving the baseline accuracy, as shown in column 8 in Tab. \ref{tab:result}.

\subsection{Ablation Study}
We conduct an ablation study to verify the effectiveness of the RL-based controller. By applying the Base-\ouralg defined in Sec. \ref{sec:setup} to the same victim models, we can measure the number of obfuscated weights required to cause the same accuracy drop, the increment ratio of the Base-\ouralg compared to \ouralg is reported in Tab. \ref{tab:ablation}.
The increment can reach up to 125\% in the worst case, demonstrating the effectiveness of the controller in optimizing filter selection. In conclusion, our developed RL-based controller achieves a drastic accuracy drop with fewer obfuscated weights.

\begin{table}[htbp]
\centering
\resizebox{0.8\linewidth}{!}{
\setlength{\tabcolsep}{15pt}
\begin{tabular}{ccc}
\toprule
Dataset & Model & Ratio                           \\
\midrule
\multirow{3}{*}{\begin{tabular}[c]{@{}c@{}}Fashion \\ MNIST\end{tabular}}  
 & VGG-11           &        
 +53.03\%                            \\ 
& ResNet-18         &         +81.82\%                                           \\ 
& MobileNet-v2       &         +26.18\%               
                 \\ \midrule
\multirow{3}{*}{CIFAR-10} 
 & VGG-11                &     +41.32\%                            \\ 
& ResNet-18              &     +81.45\%                                          \\ 
& MobileNet-v2      &       +36.05\%                                            \\ \midrule
\multirow{3}{*}{CIFAR-100} 
& VGG-11           &          +51.28\%                    \\ 
 
& ResNet-20        &                   +125.00\%                             \\ 
& MobileNet-v2     &          +89.71\%                  \\          
\bottomrule
\end{tabular}
}
\caption{The increment ratio of the obfuscated weights for Base-\ouralg compared to \ouralg when both cause the random-guessing accuracy.}
\label{tab:ablation}
\end{table}

\section{Discussion}
In addition to the effectiveness, \ouralg also considers the potential attack surfaces, i.e., whether an adversary can identify the obfuscated weights and mitigate their adverse effects, or improve the accuracy of the obfuscated model through further attacks, such as fine-tuning the model using limited training data. Thus, we evaluate the stealthiness and resilience of \ouralg, following our defined requirements in Tab. \ref{tab:requirments}.
Furthermore, we conduct a comparison between a straightforward obfuscation strategy and our method to highlight the superiority of \ouralg in terms of mitigating potential strong attacks as in Sec. \ref{sec:attack}.

\subsection{Stealthiness}
As discussed in Sec. \ref{sec:performance}, previous works achieving accuracy drop by manipulating weights fall into two categories: magnitude-based and gradient-ranking-based \cite{liu2017fault,rakin2019bit}. However, compared to the former category \cite{liu2017fault}, \ourtitle constrains the obfuscated weights within the original range of weight values, thus avoiding being easily identified.
As for the latter category, attackers can potentially locate the obfuscated weights by examining the weight gradients, allowing them to improve the degraded accuracy through weight reconstructions \cite{chen2021low}. However, \ouralg mitigates this threat by employing an optimization method instead of a greedy method. This makes it more difficult for attackers to reverse engineer the obfuscated weights and improve the accuracy based on existing knowledge, thus ensuring a high level of stealthiness.

\subsection{Resilience against Potential Attack Surfaces}
\label{sec:attack}
Following our threat model in Sec. \ref{sec:background}, we assume a strong attacker, who strives to  improve the accuracy of the obfuscated models using SOTA techniques, like norm clipping \cite{yu2021defending} and fine-tuning \cite{adi2018turning}.

\textbf{Against Norm Clipping.}
The norm clipping proposed in \cite{yu2021defending} aims to defend against universal adversarial patches by restricting the norm of feature vectors. In our case, since the accuracy drop is caused by the magnitude change of some weights (from small to large), attackers may adopt norm clipping to weights and try to clip the obfuscated weights and eliminate their adverse effect. Specifically, the weight values outside an interval will be clipped to the interval edges, where the interval is defined by
\begin{equation}
    Interval = t * [\min\{\textbf{W}+\Delta \textbf{W$'$}\}, \max\{\textbf{W}+\Delta \textbf{W$'$}\}]
\end{equation}
and t is a coefficient in the range [0, 1]. 

\begin{figure}
    \centering
    \includegraphics[width=0.47
    \textwidth]{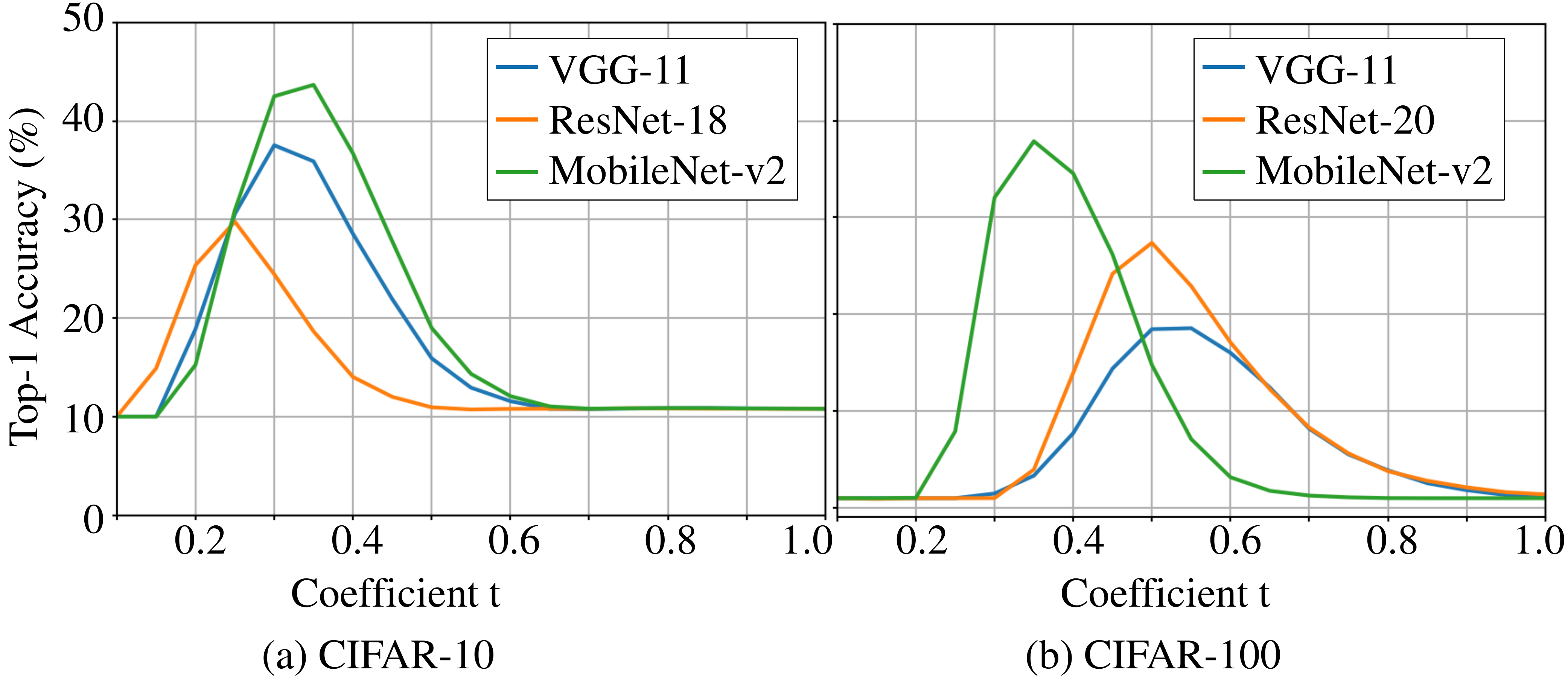}
    \caption{Apply norm clipping to improve the accuracy of obfuscated models on CIFAR-10/100.}
    \label{fig:clip}
\end{figure}

We conduct experiments to evaluate the effectiveness of norm clipping as an attack against \ouralg. The results, shown in Fig.~\ref{fig:clip}, demonstrate that as the clipping threshold decreases, the accuracy of the obfuscated models initially increases due to more obfuscated weights being clipped. However, after reaching a certain point, the accuracy starts to decrease because normal weights are also being clipped. It is important to note that  the highest  accuracy  achieved by the attacker is still below 50\%, indicating the resilience of \ouralg against norm clipping attacks.

\begin{figure}[h]
    \centering
    \includegraphics[width=0.47
    \textwidth]{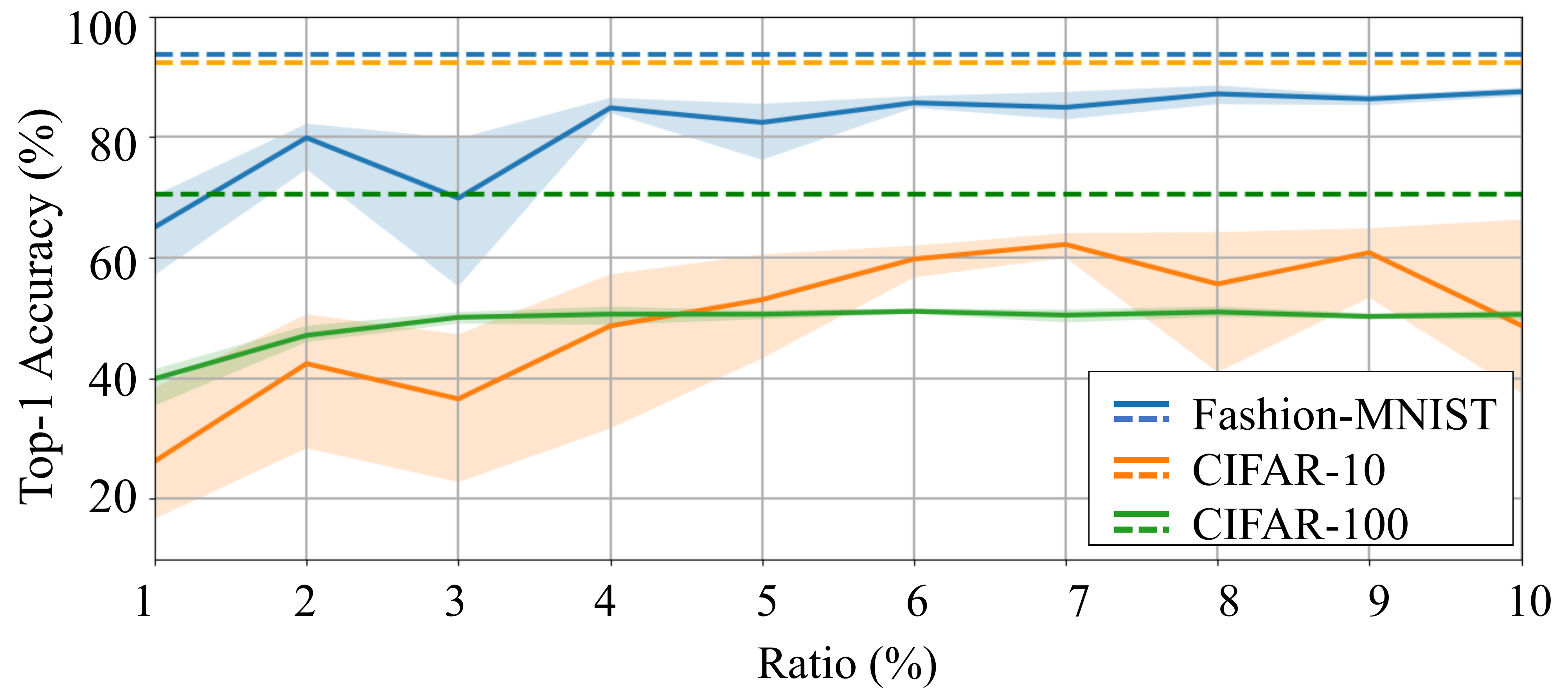}
    \caption{Apply fine-tuning to improve the accuracy of obfuscated VGG-11 models on different datasets. For each ratio, the average (solid line) and the error band (shadow region) are taken from 5 trials, with baseline accuracy as comparisons (dotted line).}
    \label{fig:fine-tuning}
\end{figure}
\textbf{Against Fine-tuning.} 
Assuming stronger attackers who are aware of weight obfuscation in various layers (as illustrated in Fig. \ref{fig:layer_dis}), they may attempt to reconstruct the weights through fine-tuning the obfuscated models using limited data \cite{adi2018turning}. To evaluate the resilience of \ouralg against fine-tuning attacks, we consider different sizes of datasets available to the attackers, ranging from 1\% to 10\% of the training data used by the victim models.
As shown in Fig. \ref{fig:fine-tuning}, in general, the accuracy will improve with the increased ratio of datasets used for fine-tuning. However, since the dataset is randomly sampled for each trial, some data may contribute more to the model fine-tuning than others, which explains the fluctuation in Fig. \ref{fig:fine-tuning}.

\begin{figure}[htbp]
    \centering
    \includegraphics[width=0.47
    \textwidth]{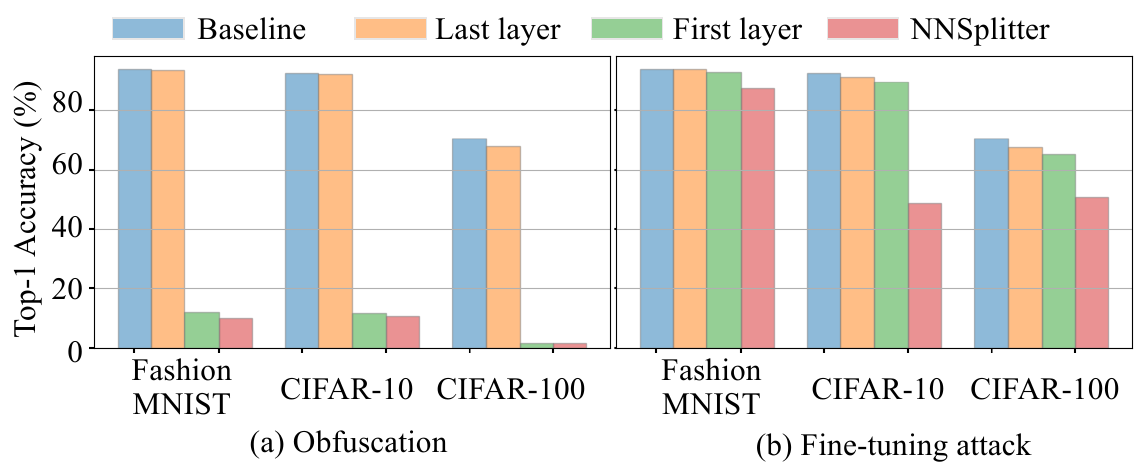}
    \caption{The performance comparison between the obfuscated model generated by \ouralg and obfuscating a single layer (either the first or the last layer). Figure (a) displays the accuracy of the obfuscated model, while (b) shows the improved accuracy achieved through fine-tuning.}
    \label{fig:fine_tuning_one}
\end{figure}

Moreover, our study demonstrates that distributing weight changes across multiple layers is more effective in protecting against the fine-tuning attack compared to concentrating them in a single layer. This finding highlights the benefit \textbf{(ii)} of our mask design as discussed in Section \ref{sec:problem}. Specifically, with the number of model secrets fixed, we add weight changes only to  the first or the last layer of VGG-11 models on three datasets, respectively, and fine-tune the obfuscated models with 10\% of training data. As shown in Fig. \ref{fig:fine_tuning_one}, obfuscating only the last layer results in a slight accuracy drop ($<$ 2\%), which could be recovered close to the baseline accuracy through the fine-tuning attack.  Although obfuscating the first layer achieves a drastic accuracy drop as \ouralg from the defense perspective, its defense effects are not resilient against the fine-tuning attack at all. In summary, our proposed \ourtitle outperforms these strategies in the expected design requirements in Tab. \ref{tab:requirments}.

\subsection{Obfuscation Strategy}
We conduct experiments using VGG-11 on CIFAR-10 to compare the straightforward obfuscation method that modifies the scale and bias parameters of the normalization layer with \ouralg. The results are presented in Tab. \ref{tab:norm}.
By obfuscating the scale parameter to 1 and the bias parameter to 0 in the normalization layer, resulting in 5504 altered parameters, the obfuscated accuracy of the model decrease significantly to 13.77\%. This demonstrates the effectiveness of the straightforward obfuscation technique in degrading the model's performance.

However, we observe that this obfuscated model is less effective in providing long-term protection against fine-tuning attacks. In particular, when attackers have access to only 10\% of the training dataset and perform fine-tuning, they are able to restore the accuracy to 59.15\%. 
In contrast, our proposed \ouralg achieves a greater accuracy drop, i.e., 10.4\% lower than obfuscating the normalization statistics, while obfuscating fewer weights (876 vs. 5504). This finding demonstrates the effectiveness of our proposed defense approach.

\begin{table}[h]
\centering
\resizebox{0.95\linewidth}{!}{
\begin{tabular}{cccc}
\toprule
& Num. / Ratio (\%) & Obfu. Acc. & Fine-tuned Acc. \\
\midrule
Scale/bias & 5504/0.019 & 13.77\% & 59.15\% \\
Weights \textbf{(our)} & \textbf{876/0.003} & \textbf{10.78\%} & \textbf{48.75\% }\\
\bottomrule
\end{tabular}}
\caption{Comparison of different obfuscation strategies.}
\label{tab:norm}
\end{table}

Furthermore, this experiment comparison verifies our intuition that reconstructing the convolutional weights is more challenging for attackers compared to reconstructing the normalization statistics, which serves as a motivation for us to design a sophisticated weight obfuscation strategy as part of our model protection approach.

\section{Conclusion}
We propose a novel model IP protection scheme \ouralg to actively protect the DNN model by preserving the model functionality  exclusively for legitimate users. 
By leveraging the support of TEE, \ouralg automatically splits a victim model into  two components: the obfuscated model, stored in the normal world, and the model secrets, securely stored in the secure world.
Through extensive experiments, we demonstrate the effectiveness of \ouralg in achieving efficient model protection, e.g., by modifying around 0.001\% weights (313 out of 28.14M), the victim model only outputs random prediction, rendering it useless for model attackers. Conversely, legitimate users can successfully execute authorized inferences by utilizing the safeguarded model secrets.
Furthermore, we address the important aspects of stealthiness and resilience against potential attacks in the design of \ouralg. This ensures that attackers are unable to identify our obfuscation technique or improve the degraded accuracy with reasonable efforts. 
By fulfilling these critical design requirements, \ouralg emerges as a promising solution for protecting DNN models in real-world scenarios. Its ability to maintain the integrity and functionality of the models while preventing attackers from unauthorized use makes it an attractive option for model owners looking to safeguard their valuable intellectual property.

\section{Acknowledgments}
This work was supported in part by the U.S. National Science Foundation under grants CNS-2247892, CNS-2153690, and CNS-2239672.

\bibliography{egbib}
\bibliographystyle{icml2023}



\end{document}